\documentclass{article}

\usepackage[T1]{fontenc}
\usepackage[utf8]{inputenc}   

\usepackage[margin=1in]{geometry}

\usepackage{graphicx}
\usepackage{subcaption}
\usepackage{booktabs}

\usepackage{tabularx}

\usepackage{url}
\usepackage{hyperref}

\usepackage[numbers]{natbib}

\usepackage{authblk}

\title{
A Scalable Pipeline Combining Procedural 3D Graphics and Guided Diffusion for Photorealistic Synthetic Training Data Generation in White Button Mushroom Segmentation
}

\author[1]{Artúr I. Károly}
\author[1]{Péter Galambos}

\affil[1]{Antal Bejczy Center for Intelligent Robotics, Research and Innovation Center of Obuda University, Budapest, Hungary}

\date{}

\begin{document}

\maketitle

\begin{abstract}

Industrial mushroom cultivation increasingly relies on computer vision for monitoring and automated harvesting. However, developing accurate detection and segmentation models requires large, precisely annotated datasets that are costly to produce. Synthetic data provides a scalable alternative, yet often lacks sufficient realism to generalize to real-world scenarios. This paper presents a novel workflow that integrates 3D rendering in Blender with a constrained diffusion model to automatically generate high-quality annotated, photorealistic synthetic images of Agaricus Bisporus mushrooms. This approach preserves full control over 3D scene configuration and annotations while achieving photorealism without the need for specialized computer graphics expertise. We release two synthetic datasets (each containing 6,000 images depicting over 250k mushroom instances) and evaluate Mask R-CNN models trained on them in a zero-shot setting. When tested on two independent real-world datasets (including a newly collected benchmark), our method achieves state-of-the-art segmentation performance (F1 = 0.859 on M18K), despite using only synthetic training data. Although the approach is demonstrated on Agaricus Bisporus mushrooms, the proposed pipeline can be readily adapted to other mushroom species or to other agricultural domains, such as fruit and leaf detection.

\end{abstract}

\section{Introduction}

In recent years, industrial mushroom cultivation for human consumption has gained growing interest, driven by their health benefits \cite{mushroom_health_benefits}, potential as a meat alternative \cite{mushroom_meat_alternative}, and overall sustainability \cite{mushroom_cultivation}. As a result, there has been a surge of research papers that focus on mushroom growth monitoring \cite{mushroom_monitoring, mushroom_monitoring_2}, cultivation optimization \cite{mushroom_cultivation_opt, mushroom_cultivation_opt_2, mushroom_cultivation_opt_3}, and automation of mushroom harvesting \cite{mushroom_harvesting, mushroom_harvesting_2, mushroom_harvesting_3}. Most of these approaches rely on computer vision and machine learning methods \cite{mushroom_cultivation_ml_review}. Usually, a key element of such approaches is the detection of mushrooms on images and/or the preparation of segmentation masks \cite{mushroom_monitoring,mushroom_monitoring_2, mushroom_cultivation_ml_review}. State-of-the-art methods usually utilize Deep Learning (DL) technologies for image-based object detection and segmentation. However, such approaches typically require extensive training datasets that are challenging and time-consuming to acquire and annotate.

In this paper, we introduce a synthetic data generation pipeline for white Agaricus Bisporus mushroom (white button mushroom) detection. The proposed method can automatically generate large and diverse photorealistic synthetic datasets along with automatically generated instance-level segmentation masks.

Before presenting the paper’s original contribution, the following subsections review related work on mushroom detection, focusing first on how existing approaches acquire training data, and then surveying current methods for generating photorealistic synthetic datasets and performing synthetic-to-real domain adaptation.

\subsection{Mushroom detection and dataset preparation}

Cong et al.~introduced a DL-based system for the detection and classification of shiitake mushrooms \cite{mushroom_detection_yolo}. Their method, MYOLO, is a refined version of the YOLOv3 model \cite{yolov3}. MYOLO improves YOLOv3’s performance for detecting objects across varying scales within a single image, as well as under conditions of severe occlusion, both of which frequently occur in images of mushrooms. Their data were acquired using an industrial camera setup, mobile phones, and web scraping. The resulting dataset comprises 1,416 images, including both close-up views and wider field-of-view shots. Annotations were created manually using LabelImg to define bounding boxes. As each image typically contains between 1 and 20 mushrooms, manual annotation remained feasible.

In \cite{mushroom_detection_mrcnn}, Wang et al.~proposed a mushroom instance segmentation network based on the Mask R-CNN model \cite{mrcnn}. They utilized a manually annotated dataset of 89 images to train their model. In our experiments, we also utilize the Mask R-CNN model to test the benefits of training on data generated by our proposed pipeline.

Yang et al.~proposed a contour-based method for accurately segmenting densely overlapping mushrooms by improving the SOLOv2 model \cite{solov2} using the PointRend module \cite{pointrend} during the up-sampling stage \cite{mushroom_detection_yolo_2_1, mushroom_detection_yolo_2_2}. Their data was acquired in a mushroom cultivation facility using three industrial cameras attached to a robotized platform. Images were captured under diverse illumination conditions, including top-, side-, and low-light setups, and they typically depict many mushrooms (more than 20). Their dataset consists of 1,920 images after data augmentation, for which they manually prepared instance-level segmentation masks using the LabelMe software. Images in their dataset closely resemble our real-world data. However, our images typically include an order of magnitude more mushrooms and capture specimens at various stages of development.

Qian et al.~utilized the Single Shot Detector (SSD) architecture for detecting oyster mushrooms in images \cite{mushroom_detection_ssd}. They used the RGB-D modalities to localize the mushrooms for robotized harvesting. Their dataset consists of 4,600 images, each depicting typically up to 10 oyster mushrooms. The images were collected over three months in multiple greenhouses at specific periods of the day to ensure consistent lighting conditions. They utilized the LabelImg software to manually annotate the images with bounding boxes.

Similarly, Retsinas et al.~also utilized depth information for mushroom localization \cite{mushroom_detection_3d_1, mushroom_detection_3d_2}. They used two RGB-D cameras to create a multi-view (9 viewpoints) 3D point cloud representation of the scene. After an initial point cloud registration step, they extracted 3D features using the Fast Point Feature Histograms (FPFH) \cite{FPFH} and Fully Convolutional Geometric Features (FCGF) \cite{FCGF} methods, followed by one-class classification on the normalized features to detect mushroom caps. They used a template 3D mushroom cap model (and its slightly altered variations for data augmentation) to train their classifiers. Afterward, they utilized the Density-Based Spatial Clustering of Applications with Noise (DBSCAN) algorithm \cite{DBSCA} to obtain instance-level segmentation. The authors highlighted that the lack of annotations was the most significant difficulty in developing and evaluating their method. As a result, they utilized synthetic mushroom scenes (containing between 5 and 45 mushrooms) to evaluate their method quantitatively, and they provided results on some real-world examples for qualitative evaluation.

\begin{figure}[t]
    \centering
    \begin{subfigure}[b]{0.3\textwidth}
        \includegraphics[width=\textwidth]{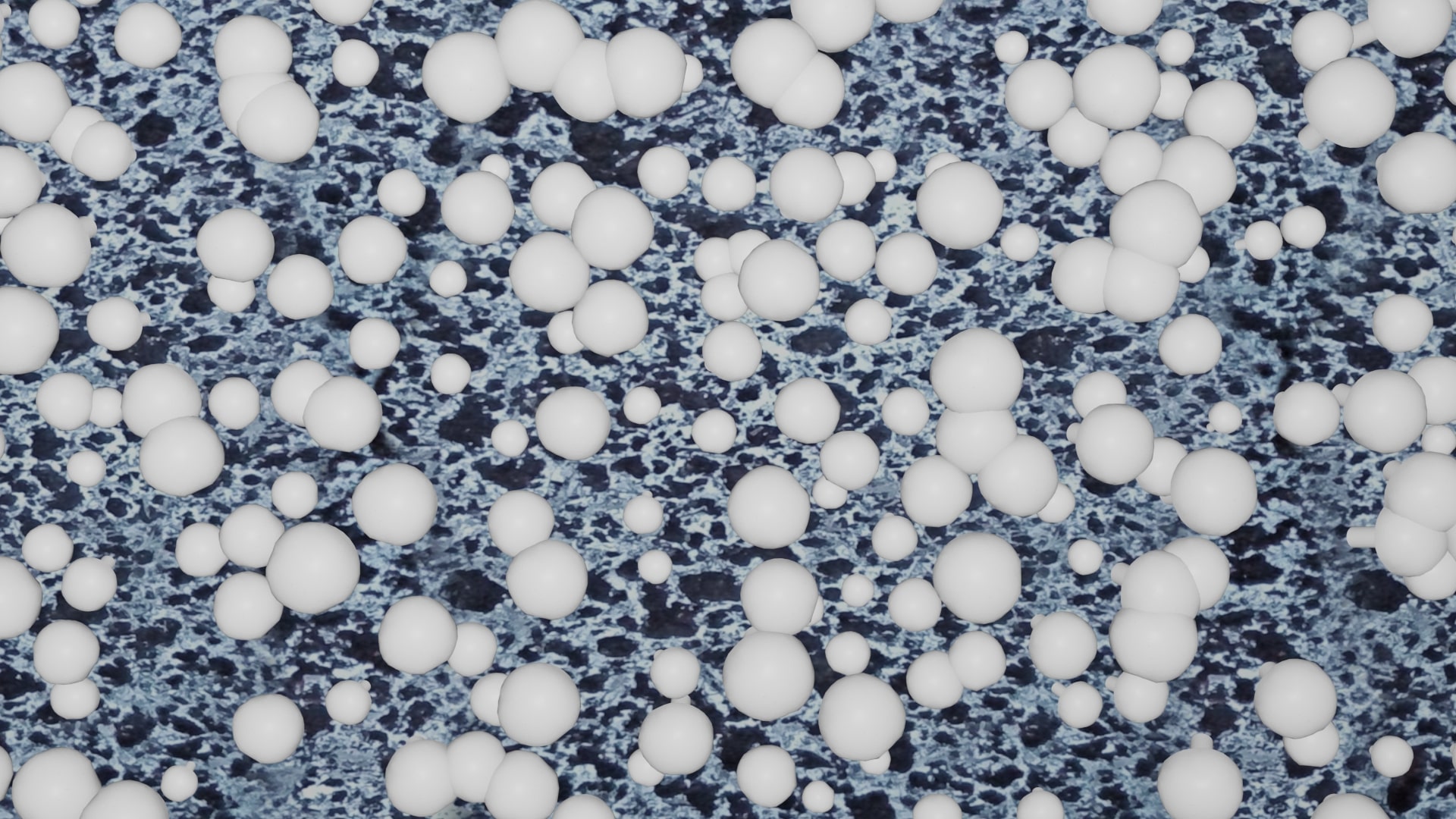}
        \caption{}
        \label{fig:subfig_1_a}
    \end{subfigure}
    \hfill
    \begin{subfigure}[b]{0.3\textwidth}
        \includegraphics[width=\textwidth]{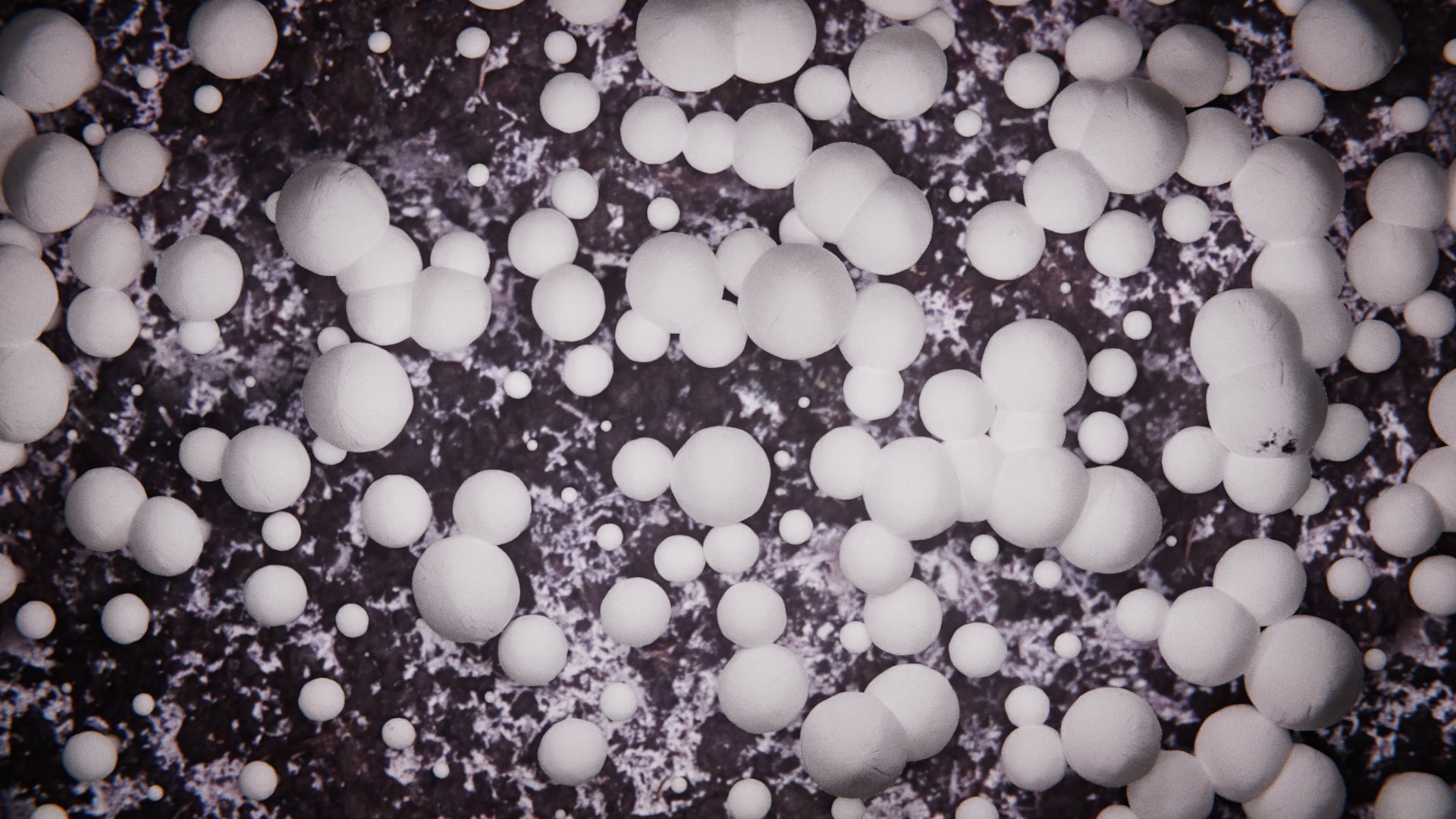}
        \caption{}
        \label{fig:subfig_1_b}
    \end{subfigure}
    \hfill
    \begin{subfigure}[b]{0.3\textwidth}
        \includegraphics[width=\textwidth]{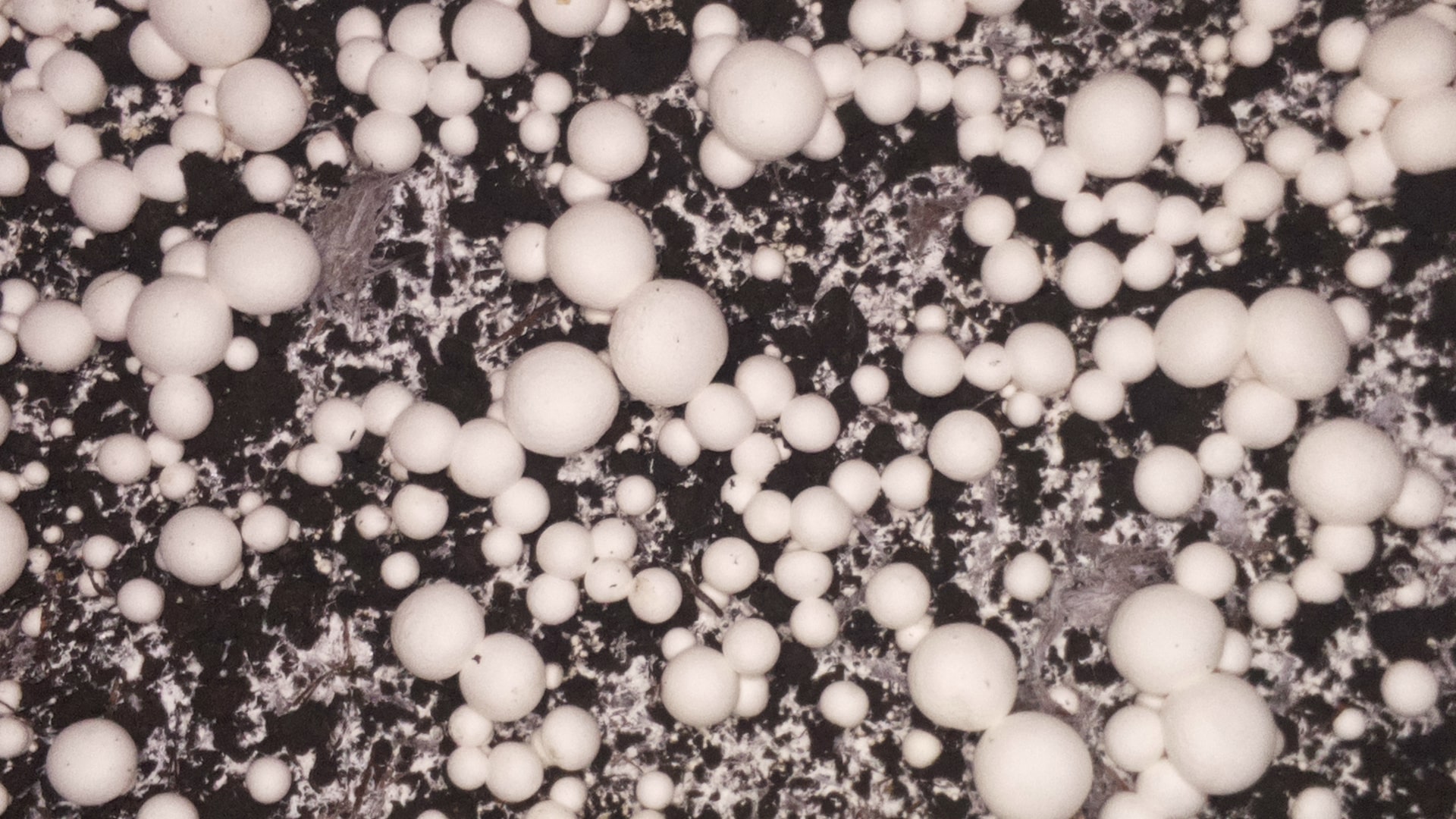}
        \caption{}
        \label{fig:subfig_1_c}
    \end{subfigure}
    \caption{Comparison of images rendered from the previous and the improved scene, and a reference real-life image. (a): Previous scene \cite{self_synthetic}. (b): Improved scene. (c): Reference real-life image.}
    \label{fig:fig1}
\end{figure}

The method proposed by Retsinas et al.~only considers 3D features while ignoring mushroom appearance. In one of our previous work, we proposed a 3D synthetic mushroom scene that considers the scene's appearance in addition to its geometry \cite{self_synthetic}. In this paper, we use an enhanced version of that scene, featuring improved realism, as the baseline for our experiments. This synthetic scene enables automated depth map and instance-level segmentation mask generation. Figure~\ref{fig:fig1}.~compares images rendered from the original and the improved synthetic scene with one of our real-life images, which was captured at an industrial mushroom farm using a Raspberry Pi-based data acquisition unit \cite{self_daq}. A clear improvement can be observed in realistic textures and lighting conditions. However, there is still a clear distinction between the synthetic images and the real-world photos. This discrepancy can cause models trained solely on synthetic data to perform poorly on real-world images. This problem is often referred to as the ``reality or domain gap'' \cite{reality_gap}, and it is usually approached either by domain randomization \cite{reality_gap} or by using photorealistic synthetic data \cite{hypersim}. In this paper, we focus on the latter approach, although we also randomize several aspects of the synthetic scene, such as soil texture, lighting conditions, mushroom distribution, as well as the number, size, and appearance of the mushrooms. 

\subsection{Photorealistic synthetic dataset preparation}

Zakeri et al.~introduced M18K, a dataset containing 423 RGB-D images of real mushrooms captured in an industrial environment using an Intel RealSense D405 camera \cite{m18k}. Each image is annotated with instance-level segmentation masks. We use the M18K dataset for comparative evaluation in our study. In their subsequent work, Zakeri et al.~developed SMS3D, a synthetic dataset designed for object detection and pose estimation tasks \cite{sms3d}. It comprises 40,000 unique RGB-D scenes, each containing up to 75 mushrooms, annotated with instance-level masks and 6D pose information. To achieve photorealism, they employed 3D scans of real mushrooms and high-resolution ground texture images. Similarly, our synthetic scenes employ textured ground planes to achieve photorealistic rendering; however, instead of relying on rigid, scan-based 3D models, we use procedurally generated mushroom geometries to enhance flexibility.

We use Blender, an open-source computer graphics software to generate our synthetic dataset \cite{blender}. One key advantage of such software is its native support for 3D environments, which allows precise control over scene geometry, composition and lighting, as well as the automatic generation of annotations like segmentation masks and depth maps. Additionally, Physically Based Rendering (PBR) enables the creation of photorealistic images. Nevertheless, achieving high realism with computer graphics tools requires specialized skills. For instance, Roberts et al.~relied on scenes crafted by professional 3D artists to produce the synthetic dataset used in Hypersim \cite{hypersim}.

Combining computer graphics with generative models can reduce the reliance on expertise in graphics software while maintaining complete control over 3D environments. This is typically accomplished through pixel-level domain adaptation, which transforms synthetic images from the computer graphics software to resemble real-world ones \cite{synth_to_real_domain_adapt}. Many of these synthetic-to-real adaptation methods \cite{gan_1, gan_2, gan_3} utilize Generative Adversarial Networks (GANs) \cite{gan}, with models like CycleGAN \cite{cyclegan}.

Liu et al.~used Blender alongside a GAN-based approach to generate a realistic RGB-D synthetic dataset aimed at robotic manipulation in cluttered environments \cite{blender_gan}. They employed 3D models with realistic textures to produce photorealistic RGB images and incorporated adversarial loss to align feature extraction between real and synthetic depth maps. Our method also uses Blender to generate synthetic RGB and depth images but emphasizes domain adaptation for RGB images, leveraging the depth modality for support. Rather than adopting an adversarial framework, we utilize text-to-image diffusion models, which provide greater flexibility than GANs and offer intuitive tools for fine-tuning outputs without requiring adversarial training.

In addition to GANs, a widely adopted unpaired image translation approach is Contrastive learning for Unpaired image-to-image Translation (CUT) \cite{cut}. Imbusch et al.~demonstrated the effectiveness of CUT for synthetic-to-real domain adaptation \cite{cut_1}, highlighting that while it may not consistently outperform CycleGAN-based methods, its straightforward objective and ease of training make it an attractive alternative. Similarly, our proposed method emphasizes simplicity and usability, leveraging the strengths of modern text-to-image diffusion models along with the efficiency of Low-Rank Adaptation (LoRA) \cite{lora}.

Zakharov et al.~proposed Photo-realistic Neural Domain Randomization (PNDR), a neural rendering-based method for synthetic-to-real domain adaptation \cite{neural_rendering}. Their approach uses a network composition that functions like a physically based ray tracer, generating high-quality images directly from scene geometry. They showed that unpaired image translation methods such as CycleGAN and CUT often distort object shapes, making them less suitable for segmentation tasks that require shape consistency. Like PNDR, our method leverages scene geometry — specifically, synthetic depth maps — while using text prompts to control appearance. This approach offers high flexibility and ensures consistent object shapes, enabling accurate and reliable segmentation mask generation for the synthesized images.

Nguyen et al.~presented Dataset Diffusion, a synthetic dataset generation method based on diffusion models \cite{dataset_diffusion}. Their approach utilizes Stable Diffusion \cite{stable_diffusion} to generate photorealistic images and corresponding semantic segmentation masks guided by text prompts. Diffusion model-based image generation is becoming increasingly prevalent in computer vision applications for modern agriculture, including tasks such as grain and fruit harvesting \cite{grain_1, grain_2, fruit}. While our method also builds on Stable Diffusion, we enhance control over image generation by incorporating depth information from the synthetic Blender scene. This setup enables precise control over parameters such as view angle, object count, size, and placement —factors that are particularly critical when training object detectors for specialized tasks like mushroom detection, where geometric variability and developmental stages strongly influence detection accuracy.

\subsection{Contributions and structure of the paper}

In summary, the key contributions of this work are:
\begin{itemize}
    \item A novel workflow for automatically generating high-quality, annotated, and photorealistic synthetic datasets for image-based instance segmentation, which we demonstrate on the Agaricus Bisporus mushroom segmentation task.
    \item Our methodology integrates 3D computer graphics with diffusion models to achieve high photorealism without requiring manual material definitions, shader configurations, or lighting setups. This hybrid approach enables implicit handling of complex visual properties while maintaining precise control over key scene parameters such as geometry and viewpoint, ensuring both realism and reproducibility in the generated data.
    \item We developed a Blender-based programmatic framework capable of generating diverse, randomized scenarios and rendering realistic images of mushrooms.
    \item We provide two datasets, each containing 6,000 images at a resolution of $1024\times1024$ with instance segmentation masks for Agaricus Bisporus mushrooms: one rendered directly from Blender and the other generated using our proposed workflow. Both datasets are publicly available on Hugging Face (ABC-iRobotics/SynWBM \cite{synwmb}).
    \item We present evaluation results of Mask R-CNN models trained on these datasets, tested against real-world data to assess their generalization performance beyond synthetic examples.
\end{itemize}

The rest of the paper is structured as follows: Section \ref{sec:mat_met} details the complete design and configuration of the Blender-based scene, including its spatial layout, object composition, camera setup, shaders, lighting, and other essential components. It also explains how the diffusion model was trained, employed, and integrated with the Blender environment to generate photorealistic synthetic images. Section \ref{sec:res} presents the characteristics of the generated datasets, experimental design, and evaluation results, complemented by an ablation study. Section \ref{sec:disc} analyzes the strengths and limitations of the proposed method, discussing failure cases, generalization performance, and directions for future research. Finally, Section \ref{sec:conc} concludes the paper by summarizing the key findings and contributions.

\section{Materials and Methods}
\label{sec:mat_met}

\subsection{Blender scene}

The central element of our synthetic data generation pipeline is the synthetic scene constructed in Blender. The scene geometry includes mushroom instances and a ground plane. To ensure flexibility, we employed Blender's Geometry Nodes to design a procedural mushroom model. This model features a closed cap (gills are not visible) with randomized stem length and cap shape. Parameters of the model include

\begin{itemize}
    \item \textbf{Age}: Represents the mushroom’s age, directly influencing its overall size. This parameter ranges from $0.05$ to $1$. An age of 0.05 corresponds to a cap diameter of 1.5 mm, while an age of 1 represents a cap diameter of 30 mm.
    \item \textbf{Age increment}: Used for generating multiple instances of a mushroom model at varying ages. While it does not affect appearance directly, it enables automated age variation by chaining model node groups. The incremented age is output by the node group, facilitating seamless chaining. Valid values lie within $[0, 1]$.
    \item \textbf{Randomness}: Controls the degree of variation in stem length and cap shape. A value of 0 applies no variation, while 1 allows maximum deviation from nominal dimensions. These deviations are relative to the model’s base size and were empirically defined.
    \item \textbf{Seed}: Sets the random seed for reproducible variation. Mushrooms with the same seed and parameters will have identical shapes.
\end{itemize}

\begin{figure}[t]
    \centering

    \begin{subfigure}[b]{\textwidth}
        \centering
        \includegraphics[width=\textwidth]{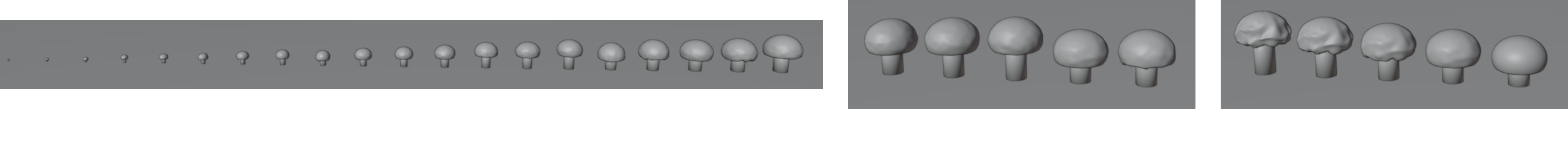}
        \caption{}
        \label{fig:subfig_2_a}
    \end{subfigure}

    \vspace{1em}

    \begin{subfigure}[b]{\textwidth}
        \centering
        \includegraphics[width=\textwidth]{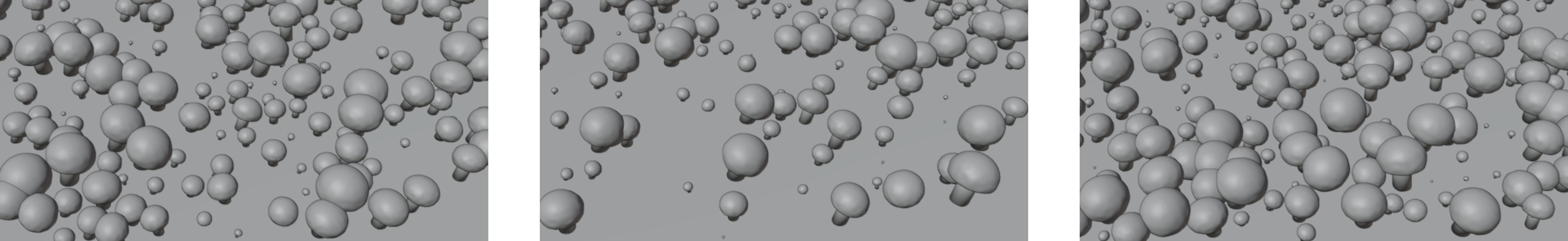}
        \caption{}
        \label{fig:subfig_2_b}
    \end{subfigure}

    \caption{Procedural mushroom models and mushroom distribution. (a): From left to right: mushrooms of varying ages, random seeds, and randomness. (b): Sample distributions of mushrooms within the synthetic scene.}
    \label{fig:fig2}
\end{figure}

Figure~\ref{fig:fig2}/a illustrates examples of procedurally generated mushroom models with varying age, random seed values, and randomness.

For the synthetic scene, we generate four rows of mushrooms, each containing mushrooms at different growth stages (as in the first image of Figure~\ref{fig:fig2}/a). Each row is initialized with a unique random seed, and individual mushrooms are assigned a randomness value of 0.25. This results in 80 distinct mushroom model variations. From these, we randomly select instances and place them at randomly distributed points on the ground plane applying random rotation around the vertical axis (ranging from 0 to 360 degrees) and random tilt around the horizontal axes (ranging from -10 to 10 degrees). The point distribution on the ground plane is guided by Poisson Disk sampling \cite{poisson_disk} with empirically determined parameters. Examples of randomly distributed mushroom models on the ground plane using this method are shown in Figure~\ref{fig:fig2}/b. As seen in the figure, the Poisson Disk sampling combined with the random selection of instances results in mushroom distributions that appear natural with a high degree of diversity in terms of size, shape, density, and orientation. The natural clustering present in our synthetic dataset often results in mushrooms being heavily occluded within the images, which in turn promotes robustness to occlusion in the models trained on this data.

For rendering photorealistic images, the mushrooms are assigned a wooden shader (eucalyptus macarthurii) with a white base color, sourced from the BlenderKit library \cite{blenderkit}. The ground plane is textured with three images obtained from our real-world data collection setup, showing the mushroom compost before the mushrooms began to grow. During rendering, one of the three available image textures is randomly selected and applied. For each texture, we generated displacement maps using the DeepBump Blender addon \cite{deepbump} to ensure proper light interaction with the ground plane. Additionally, the image texture is randomly shifted along the horizontal axes for each render.

The lighting setup includes subtle, uniform illumination from a gray High Dynamic Range Image (HDRI), a 6~W area light positioned 0.6~m above the ground plane, and eight 100~mW point light sources placed 0.15~m above the ground plane. The color of each light is randomly varied for each render.

\begin{figure}[t]
    \centering
    \includegraphics[width=0.8\columnwidth]{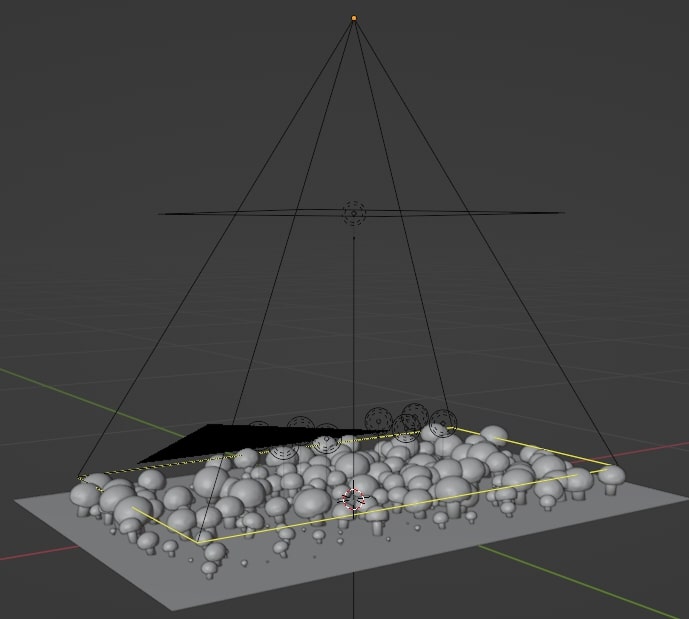}
    \caption{Synthetic scene layout with mushrooms, ground plane, lights, and camera}
    \label{fig:fig3}
\end{figure}

The camera in the scene is positioned 1~m above the ground plane facing directly downward at the mushrooms. It uses the standard Blender camera settings. During post-processing, camera lens distortion, glare, video noise, light leaks, and vignetting are added to the rendered images. An example of a synthetic image rendered from this scene is shown in Figure~\ref{fig:fig1}/b. The full scene layout is illustrated in Figure~\ref{fig:fig3}.

We used the Blender Annotation Tool (BAT) \cite{self_synthetic} to automatically generate depth data and instance-level segmentation masks corresponding to the rendered images. The segmentation masks serve as the ground-truth for training Mask R-CNN models for mushroom segmentation, while the depth data is utilized as geometric constrain to guide the image generation process with Stable Diffusion. Rendering was performed on an NVIDIA RTX 2000 Ada graphics card, requiring approximately 12 seconds to produce a single frame along with its corresponding annotations.

\subsection{Diffusion models for photorealistic image generation}

Creating photorealistic renders in Blender is challenging because achieving realism requires careful manual configuration of multiple interdependent components—materials, shaders, lighting, and camera effects. While modeling scene geometry is relatively straightforward, producing realistic lighting balance, physically based shader networks, and natural imperfections necessitates extensive expertise in computer graphics.

In contrast, text-to-image diffusion models such as Stable Diffusion \cite{stable_diffusion} can automatically generate realistic lighting, textures, and global illumination effects without manual setup. By learning these relationships directly from large-scale image datasets, they bypass the need for technical rendering knowledge, making the process substantially faster and more accessible while still yielding high-quality photorealistic outputs.

\begin{figure}[t]
    \centering
    \includegraphics[width=\textwidth]{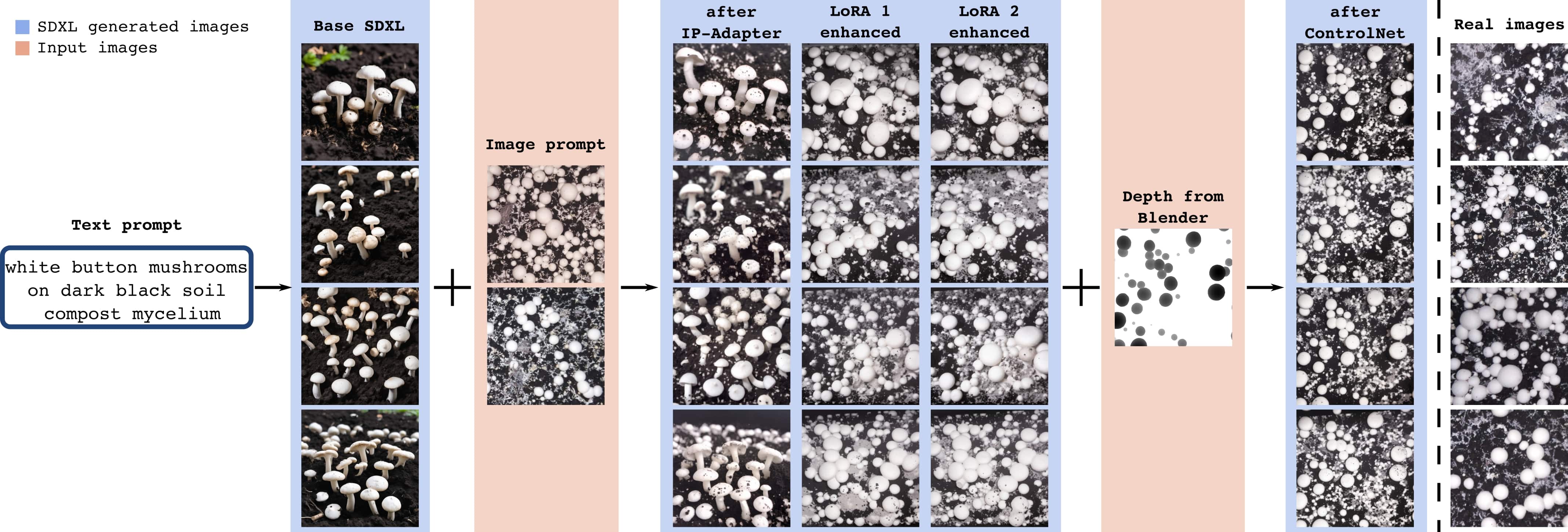}
    \caption{Diffusion model-based image generation workflow.}
    \label{fig:fig4}
\end{figure}

Our synthetic data generation pipeline employs the Stable Diffusion XL (SDXL) model \cite{sdxl} through ComfyUI \cite{comfyui}, a node-based interface that enables the creation of modular, block-oriented image generation workflows. Despite its flexibility, guiding the base SDXL model to produce images that closely match our on-site photographs (e.g., Figure~\ref{fig:fig1}/c) remains challenging, as the model tends to generate outputs with stylistic or contextual deviations from real-world scenes. The first column in Figure~\ref{fig:fig4}.~illustrates images synthesized by the base SDXL model using a simple text prompt. As seen, these images are still quite different from the real images shown in Figure~\ref{fig:fig1}/c and the last column of Figure~\ref{fig:fig4}.

To adapt the SDXL model to our specific domain, we first apply the Image Prompt Adapter (IP-Adapter) \cite{ip_adapter}, which allows the use of images as prompts by leveraging a pre-trained image encoder (CLIP \cite{clip}) and linking image features to Stable Diffusion’s U-Net via decoupled cross-attention. This enables image-based style transfer, which we use to steer the generation process toward reproducing camera effects and color schemes that resemble our real images. We selected two images as style references, shown in the second column of Figure~\ref{fig:fig4}., while the third column displays the synthesized outputs after applying the IP-Adapter.

To further enhance image generation quality, we fine-tuned the SDXL model using the Low-Rank Adaptation (LoRA) technique \cite{lora} to more accurately depict white button mushrooms, compost, and mycelium. Two LoRA models were trained: one focused on white button mushrooms and the other on compost and mycelium. The training was conducted using CivitAI’s LoRA trainer service utilizing 15 images for the first LoRA (2 from our real dataset and the rest sourced from the internet) and 12 images for the second (4 from our real dataset and the remainder also scraped online). Text descriptions for all training images were created manually. Each description included keywords intended to activate the LoRA modules during inference, specifically, ``white button mushroom'' and ``compost mycelium''. Any additional visual elements present in the training images that were not intended to be captured by the LoRA (e.g., containers or people) were also explicitly mentioned in the text prompts to help disentangle them from the target concepts (e.g., ``man holding white button mushrooms'' or ``sack of compost''). Training was conducted using the Kohya-SS backend \cite{kohya} for 10 epochs with 200 repeats, a batch size of 4, a U-Net learning rate of 0.0005, a text encoder learning rate of 0.00005, and the Adafactor optimizer paired with a cosine-with-restarts learning rate scheduler. The default mean squared error (MSE) loss function was used during training. Since the trainer service utilizes cloud resources, the exact training hardware is unknown; however, training for 10 epochs took less than 30 minutes. The impact of these LoRA models on image generation is illustrated in the fourth and fifth columns of Figure~\ref{fig:fig4}.

Finally, we incorporate a depth-based ControlNet \cite{controlnet}, which enables constrained image generation using depth maps extracted from the synthetic Blender scene. This ensures geometric consistency between the segmentation masks produced in Blender and the corresponding generated images. An example depth map and the corresponding generated images after applying ControlNet are shown in the sixth and seventh columns of Figure~\ref{fig:fig4}. The final column presents randomly selected real images allowing for a qualitative comparison of image quality and domain similarity. Notably, although each row in the figure starts from a different random seed, most of the variation in the resulting images appears in the background, since ControlNet constrains mushroom placement according to the geometry encoded in the depth map. Varying the random seed could be a useful data augmentation technique, though it becomes less critical given the abundance of unique depth maps we can produce.

This diffusion model-based workflow enables the generation of photorealistic images for our synthetic scenes without the need to manually configure lighting, shaders, or textures in Blender—a process that is both time-intensive and demands specialized expertise. Leveraging GPU acceleration with an NVIDIA RTX 2000 Ada graphics card, a single image can be generated in approximately 45 seconds.

In our experiments, models trained on images produced through this workflow outperform those trained on photorealistic renders created using manually designed shaders and lighting setups, demonstrating the efficiency and effectiveness of the proposed approach.

\section{Results}
\label{sec:res}

\subsection{Dataset and Experimental Setup}

In our experiments, we compare the performance of two Mask R-CNN models: one trained on photorealistic images rendered directly from our Blender scene (MRCNN-B) and another trained on images generated using the proposed Stable Diffusion-based pipeline (MRCNN-SD). This comparison allows us to assess the effectiveness of diffusion-based synthetic data generation relative to traditional rendering approaches for mushroom instance segmentation tasks. The evaluation is conducted on 195 real-world images from our on-field setup, which were manually annotated using Label Studio \cite{label_studio}. Additionally, we compare the zero-shot performance of both models against other mushroom segmentation methods using images from the M18K dataset \cite{m18k}.

From our Blender scene, we rendered a total of \textbf{2,000} unique photorealistic images, each accompanied by instance-level segmentation masks and depth maps. The images have a resolution of $1920 \times 1080$ pixels and collectively include \textbf{289,930} mushroom instances, resulting in an \textbf{average of 145 instances per image}.

We cropped three smaller regions ($1024 \times 1024$ pixels) from each rendered image, resulting in a final dataset of \textbf{6,000} rendered images, a subset of which was used to train the MRCNN-B model. Likewise, each corresponding depth map was cropped and used as input for the depth ControlNet in the Stable Diffusion-based generation pipeline, producing \textbf{6,000} generated photorealistic images, some of which were used to train the MRCNN-SD model. Both models were trained with the Detectron2 framework \cite{detectron2} on an NVIDIA RTX 2000 Ada graphics card while following the same hyperparameters, training strategy, and segmentation masks. The training was initialized with COCO pre-trained weights, using a batch size of 3 and a base learning rate of 0.00025. Each model was trained on 1,200 images for 50 epochs, with validation conducted on 30 images at the end of each epoch. Training a single model took approximately 8 to 9 hours. The best-performing model weights, which we used in our experiments, were selected based on the F1 score achieved during validation.

In our first experiment, we evaluated the zero-shot performance of the MRCNN-B and MRCNN-SD models using \textbf{195} manually labeled images from our real-world dataset. Each of these $512 \times 512$ resolution images were cropped from \textbf{13} high-resolution ($1920 \times 1080$) images that represent diverse conditions including varying locations, lighting, mushroom densities, and developmental stages, captured on different dates. Cropping was performed via the SAHI framework \cite{sahi} with a 20\% overlap. The high-resolution images were captured at an industrial mushroom harvesting site using Raspberry Pi-based data collection units with Raspberry Pi cameras \cite{self_daq}. In total, they contain \textbf{2,224} mushroom instances, averaging \textbf{over 170 mushrooms per image}. In comparison to the test splits of recent real-world mushroom detection datasets \cite{mushroom_detection_mrcnn, mushroom_detection_yolo_2_1, mushroom_detection_yolo_2_2}, and in particular, the M18K dataset, our test set features a higher number of mushroom instances and greater diversity in scene composition.

In our second experiment, we assessed the zero-shot performance of the MRCNN-B and MRCNN-SD models on the M18K dataset \cite{m18k} and compared their results with models trained directly on M18K. Evaluation was conducted using all those images from the test split of the M18K dataset that feature white button mushrooms. This subset comprises \textbf{20} images containing a total of \textbf{453} mushroom instances, averaging \textbf{23 mushrooms per image}. This experiment underscores the transferability of models trained on datasets generated using our approach, showcasing the high quality and generalization potential of the proposed synthetic data generation pipeline.

\subsection{Model Comparison}
For model comparison, we report standard instance segmentation metrics: Average Precision (AP), Average Recall (AR), F1 score, and mean Intersection over Union (mIoU). The results of the first experiment, comparing the MRCNN-B and MRCNN-SD models on our real-world dataset, are presented in Table \ref{tab:results1}.

\begin{table}[t]
    \centering
    \caption{Zero-shot performance of MRCNN-B and MRCNN-SD on real-world data}
    \begin{tabularx}{\textwidth}{
    >{\raggedright\arraybackslash}X
    >{\centering\arraybackslash}X
    >{\centering\arraybackslash}X
    >{\centering\arraybackslash}X
    >{\centering\arraybackslash}X
    }
        \toprule
       Model & F1 & AP & AR & mIoU\\
       \midrule
       MRCNN-B & \textbf{0.858} & 0.926 & \textbf{0.799} & \textbf{0.900}\\
       MRCNN-SD & 0.810 & \textbf{0.941} & 0.704 & 0.893\\
       \bottomrule
    \end{tabularx}
    \label{tab:results1}
\end{table}

Table \ref{tab:results1}.~indicates that the MRCNN-B model slightly outperformed the MRCNN-SD model, primarily due to a lower Recall observed in MRCNN-SD. This can be attributed to the fact that, while ControlNet provides guidance during the SDXL-based image generation process, it does not strictly prevent mushroom placement on flat regions of the depth map. As a result, the pipeline tends to produce images with a higher number of small mushrooms, many of which do not have corresponding ground-truth masks. This leads models trained on such data to favor Precision over Recall, which is reflected in the results. Despite this tendency, the MRCNN-SD model offers a significant advantage in data preparation effort. Unlike MRCNN-B, the MRCNN-SD pipeline requires no manual setup of the Blender scene for photorealistic rendering.

The results of our second experiment are shown in Table \ref{tab:results2}.,~where we compare the zero-shot performance of the MRCNN-B and MRCNN-SD models to other segmentation models trained directly on the M18K dataset. Baseline resluts are taken from the M18K paper \cite{m18k}. Since the original study did not report mean Intersection over Union (mIoU) scores, the table includes only Average Precision (AP), Average Recall (AR), and F1 metrics.

\begin{table}[t]
    \centering
    \caption{Evaluating zero-shot performance of MRCNN-B and MRCNN-SD against M18K-trained segmentation models. In each column, the highest score is highlighted using bold and underlined text, while the second-highest score is highlighted using bold text only.}
    \begin{tabularx}{\textwidth}{
    >{\raggedright\arraybackslash}X
    >{\centering\arraybackslash}X
    >{\centering\arraybackslash}X
    >{\centering\arraybackslash}X
    }
        \toprule
       Model name & F1 & AP & AR\\
       \midrule
       Mask~R-CNN & \textbf{\underline{0.881}} &  \textbf{\underline{0.866}} & \textbf{\underline{0.896}} \\
       YOLOV8-L &  0.857 & 0.846 & 0.868 \\
       YOLOV8-M & 0.844 & 0.833 & 0.856 \\
       YOLOV8-N & \textbf{0.859} & 0.846 & 0.874 \\
       YOLOV8-S &  0.857 & 0.845 & 0.869 \\
       YOLOV8-X & 0.858 & \textbf{0.847} & 0.869 \\
       MRCNN-B & 0.848 & 0.814 & \textbf{0.886} \\
       MRCNN-SD & \textbf{0.859} & 0.843 & 0.876 \\
       \bottomrule
    \end{tabularx}
    \label{tab:results2}
\end{table}

Table \ref{tab:results2} shows that both the MRCNN-B and MRCNN-SD models achieved performance comparable to segmentation models trained directly on the M18K dataset, despite being trained exclusively on synthetic data. While their Average Recall (AR) ranks among the highest, their relatively lower Average Precision (AP) can be attributed to incomplete labeling in the M18K test set, where several correctly detected mushrooms were absent from the ground-truth annotations, resulting in apparent false positives. Nonetheless, these findings demonstrate the robustness of the proposed synthetic data generation pipeline, highlighting its potential to substantially reduce manual data collection and annotation efforts without compromising model accuracy.

It is worth noting that the more conservative detection behavior of the MRCNN-SD model enabled it to outperform MRCNN-B on the M18K dataset. This advantage likely stems from the fact that, unlike our real-world dataset, the M18K images include numerous small mushroom pins, which the MRCNN-SD model tends to disregard as background rather than incorrectly classifying as instances. This selective sensitivity contributes to its superior performance in datasets where small or ambiguous objects are less reliably annotated.

\begin{figure}[t]
    \centering

    \begin{subfigure}[b]{0.6\textwidth}
        \centering
        \includegraphics[width=\textwidth]{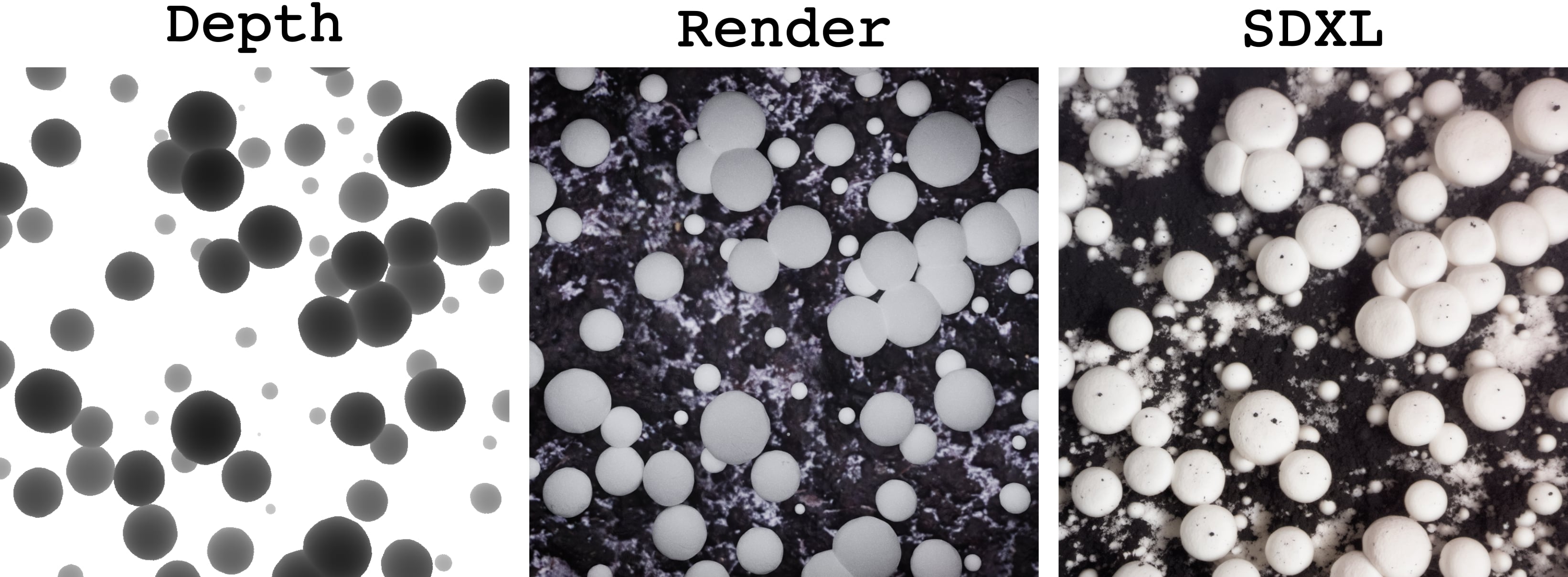}
        \caption{}
        \label{fig:subfig_5_a}
    \end{subfigure}

    \vspace{1em}

    \begin{subfigure}[b]{\textwidth}
        \centering
        \includegraphics[width=\textwidth]{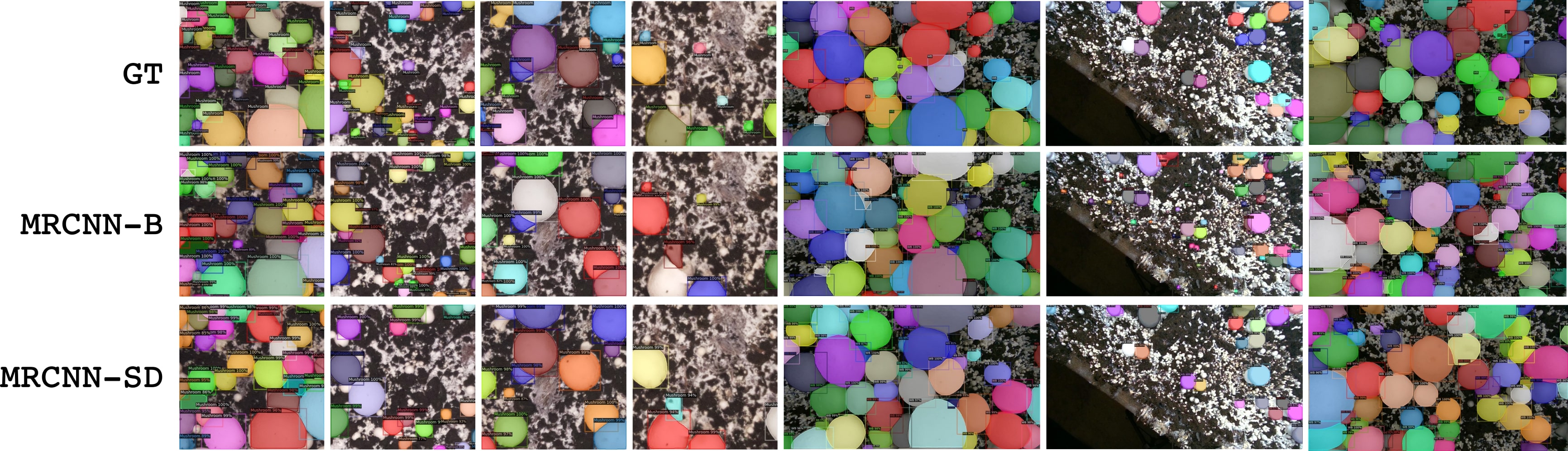}
        \caption{}
        \label{fig:subfig_5_b}
    \end{subfigure}

    \caption{Results for qualitative evaluation. (a): Comparison of rendered and generated images of mushrooms. (b): MRCNN-B and MRCNN-SD model outputs compared to the ground-truth (GT) masks.}
    \label{fig:fig5}
\end{figure}

Figure \ref{fig:fig5} presents a qualitative comparison of generated images alongside segmentation outputs from the MRCNN-B and MRCNN-SD models, compared with ground-truth (GT) annotations. The comparison includes four representative images from our real-world dataset and three images from the M18K test set.

\subsection{Ablation study}

To quantitatively assess the realism of our generated images, we compute the Fréchet Inception Distance (FID) \cite{fid} and Kernel Inception Distance (KID) \cite{kid} between the real-world and synthetic image sets using the torch\_fidelity tool \cite{torchfidelity}. Using these metrics, we perform a focused ablation study, supported by both quantitative and qualitative results, to elucidate the contributions of key components in our workflow, including the IP-Adapter, LoRAs, and ControlNet.

To establish a baseline, we first split our real-world dataset in half and computed FID and KID scores between the two halves (referred to as SAME in Table~\ref{tab:results3}.), capturing intra-domain variance and providing a lower bound for expected metric values. We then compared our real-world data to white button mushroom images from the M18K training split to quantify the domain gap between real-world datasets. Using this as a reference, we report FID and KID scores for 100 images each from our Blender-rendered and Stable Diffusion-generated datasets, relative to our real images. Table~\ref{tab:results3}.~summarizes the scores for each split compared to the first half of our real-world dataset.

\begin{table}[t]
    \centering
    \caption{Quantitative evaluation of dataset similarity relative to our real-world dataset. ``SAME'' was measured by comparing two halves of our real-world dataset, while all other datasets were compared to the first half. For KID, both the mean and standard deviation are reported. A downward arrow indicates that lower values correspond to greater similarity.}
    \begin{tabularx}{\textwidth}{
    >{\raggedright\arraybackslash}X
    >{\centering\arraybackslash}X
    >{\centering\arraybackslash}X
    >{\centering\arraybackslash}X
    }
        \toprule
       Data source & FID$\downarrow$ & KID mean$\downarrow$ & KID std\\
       \midrule
       SAME & 72.88 & 0.0094 & 0.0050\\
       M18K & 154.70 & 0.1138 & 0.0058\\
       SDXL & 235.86 & 0.2976 & 0.0172\\
       Blender & 373.82 & 0.6841 & 0.0310\\
       \bottomrule
    \end{tabularx}
    \label{tab:results3}
\end{table}

As expected, the same-domain split of our real-world dataset yields the lowest FID and KID scores, confirming that the metrics effectively capture dataset similarity and thus realism. The relatively high FID within this split also reflects the dataset’s high diversity. The next best scores come from the M18K training split, which aligns with expectations, as the domain gap between the two real datasets is smaller than that between real and synthetic data. The FID and KID values obtained for the M18K dataset serve as a benchmark. When synthetic data achieves comparable scores, it indicates that the generated images are visually indistinguishable from real-world data. While SDXL-generated images have not yet reached this threshold, they show substantially better similarity than Blender-rendered images, supporting our claim that the proposed workflow improves realism while streamlining dataset creation.

For the ablation study, we prepare the following datasets:
\begin{itemize}
    \item \textbf{BSD}: Images generated using the base SDXL model with only text prompt guidance. These do not align with Blender-generated segmentation masks (due to the absence of ControlNet) and are not usable for training, but serve as a baseline to assess visual similarity to real-world data.
    \item \textbf{+CNET}: Images generated using SDXL with ControlNet guidance.
    \item \textbf{+IP}: SDXL with ControlNet and the IP-Adapter.
    \item \textbf{+L1}: SDXL with ControlNet and the first LoRA trained on white button mushrooms.
    \item \textbf{+L2}: SDXL with ControlNet and the second LoRA trained on compost and mycelium.
    \item \textbf{+IP+L1}: All components except the second LoRA (compost and mycelium).
    \item \textbf{+IP+L2}: All components except the first LoRA (white button mushrooms).
    \item \textbf{+L1+L2}: All components except the IP-Adapter.
\end{itemize}

For each dataset, we generated 100 images and compared them to the same real-world split used in Table~\ref{tab:results3}. All images were created using the same text (and image) prompts, set of random seeds, and depth maps (when applicable), ensuring that any visual differences arise solely from the inclusion or omission of specific components. Table~\ref{tab:results4}.~presents the corresponding FID and KID scores.

\begin{table}[t]
    \centering
    \caption{Quantitative ablation study showing dataset similarity to our real-world dataset when excluding different components of the proposed workflow.}
    \begin{tabularx}{\textwidth}{
    >{\raggedright\arraybackslash}X
    >{\centering\arraybackslash}X
    >{\centering\arraybackslash}X
    >{\centering\arraybackslash}X
    }
        \toprule
       Data source & FID$\downarrow$ & KID mean$\downarrow$ & KID std\\
       \midrule
       BSD & 436.25 & 0.6175 & 0.0199\\
       +CNET & 276.38 & 0.2863 & 0.0118\\
       +IP & 295.79 & 0.2988 & 0.0183\\
       +L1 & 235.82 & 0.2621 & 0.0119\\
       +L2 & 253.80 & 0.2711 & 0.0106\\
       +IP+L1 & 234.35 & 0.2687 & 0.0138\\
       +IP+L2 & 273.31 & 0.2626 & 0.0177\\
       +L1+L2 & 244.77 & 0.2803 & 0.0129\\
       \bottomrule
    \end{tabularx}
    \label{tab:results4}
\end{table}

\begin{figure}[t]
    \centering
    \includegraphics[width=\textwidth]{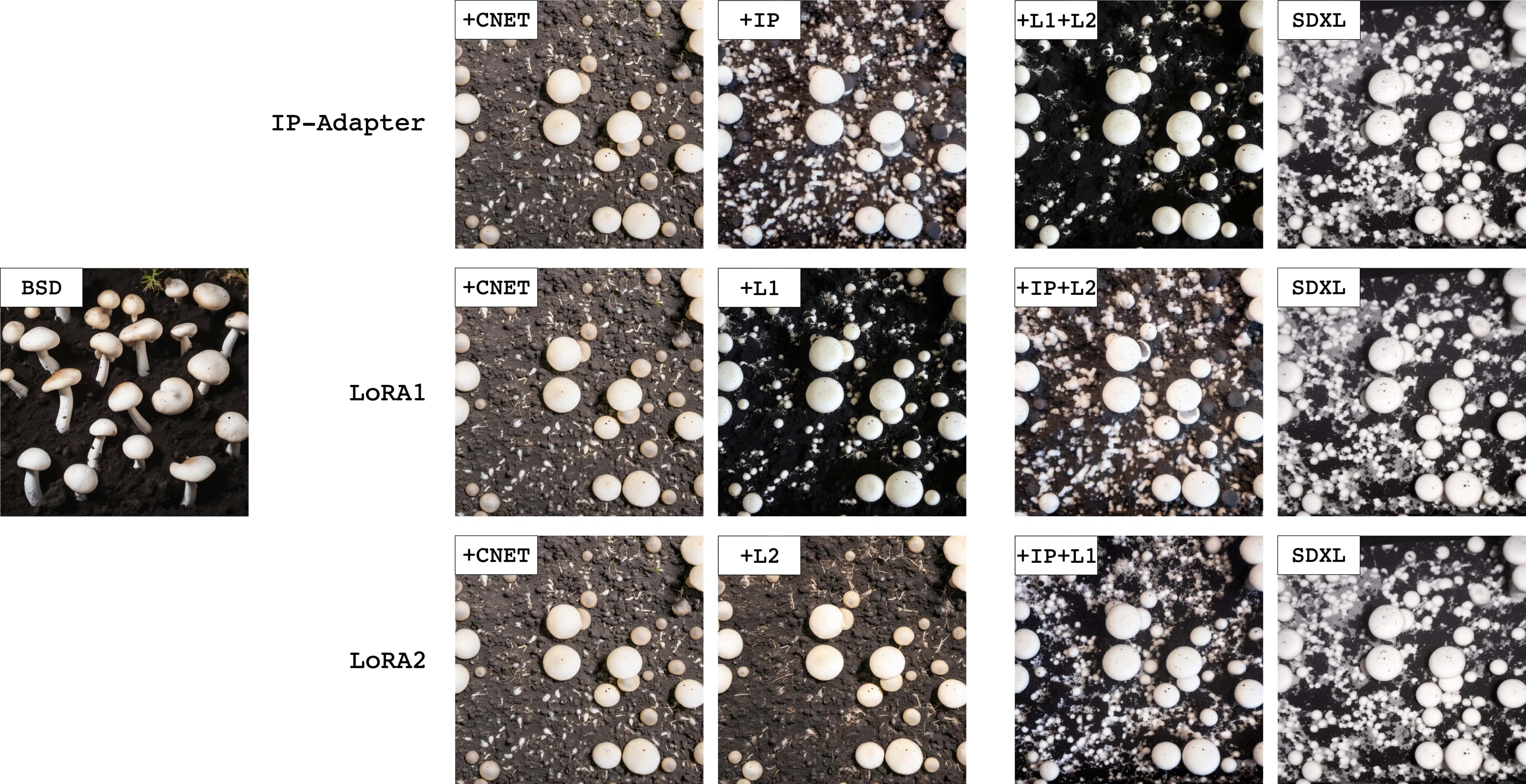}
    \caption{Qualitative ablation study showing the effects of adding or removing key components in the proposed workflow.}
    \label{fig:fig6}
\end{figure}

Figure~\ref{fig:fig6}.~complements the quantitative ablation results with a qualitative comparison of generated images. All images were created using the same random seeds, text prompts, image prompts (if applicable), and depth maps (if applicable). On the left, the image generated using the base SDXL model serves as a reference. The three rows illustrate the effects of the IP-Adapter, LoRA 1, and LoRA 2, respectively. In each row, the first image pair shows the impact of adding the component to the ControlNet-guided SDXL model, while the second pair shows the effect of removing that component from the full workflow (comparing results without the IP-Adapter/LoRA to those with all components enabled).

The images vividly demonstrate the distinct contributions of each component in the workflow. ControlNet ensures that mushroom placement aligns with the depth map (and consequently the segmentations). Due to its effect, mushrooms consistently appear in the same locations across ControlNet-guided images. The IP-Adapter controls overall image style (color, blur, etc.), LoRA1 enhances mushroom realism, while LoRA2 adjusts the background, soil, and compost.

The most realistic images (+L1, +L1+L2, +IP+L1, SDXL) also achieved the best FID and KID scores. Notably, all top-performing combinations include LoRA1, highlighting its importance alongside ControlNet. This consistency aligns with its contribution to enhancing the photorealism and visual fidelity of the mushrooms, the primary subject of the generated images.

It is important to note that FID and KID scores do not always reflect model performance; for example, although the Blender-rendered dataset scored worse, the MRCNN-B model did not perform proportionally worse than MRCNN-SD. Additionally, some component combinations (+L1, +IP+L1) had similar or better FID and KID scores than the full pipeline (SDXL), yet human evaluation preferred the images from SDXL. This aligns with known limitations of FID and KID, which do not always correlate with human judgment \cite{fid_problems}. While these metrics are useful for assessing overall image quality---such as demonstrating that the domain gap between images generated by our workflow and real-world data is smaller than that of Blender-rendered images---they have inherent limitations and should not be solely relied upon for detailed comparisons between methods.

\section{Discussion}
\label{sec:disc}

\subsection{Capabilities and challenges}

\begin{figure}[!ht]
    \centering
    \begin{subfigure}[b]{0.3\textwidth}
        \includegraphics[width=\textwidth]{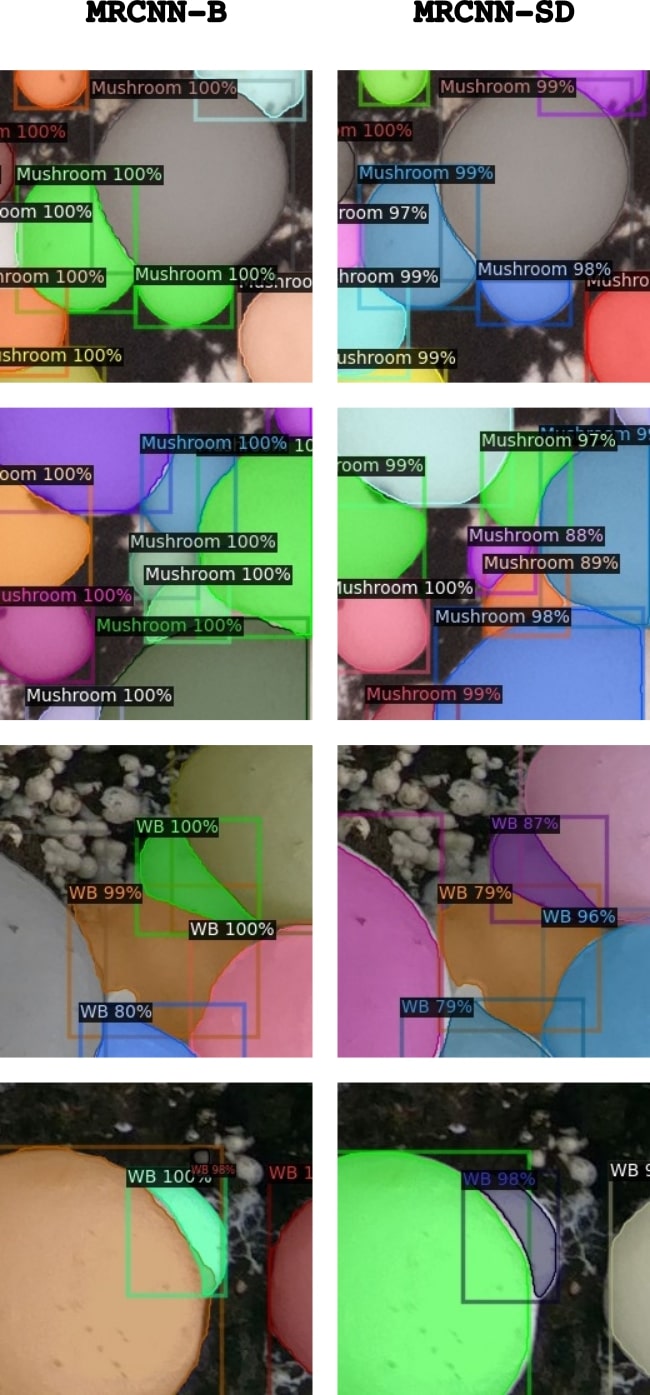}
        \caption{}
        \label{fig:subfig_7_a}
    \end{subfigure}
    \hfill
    \begin{subfigure}[b]{0.3\textwidth}
        \includegraphics[width=\textwidth]{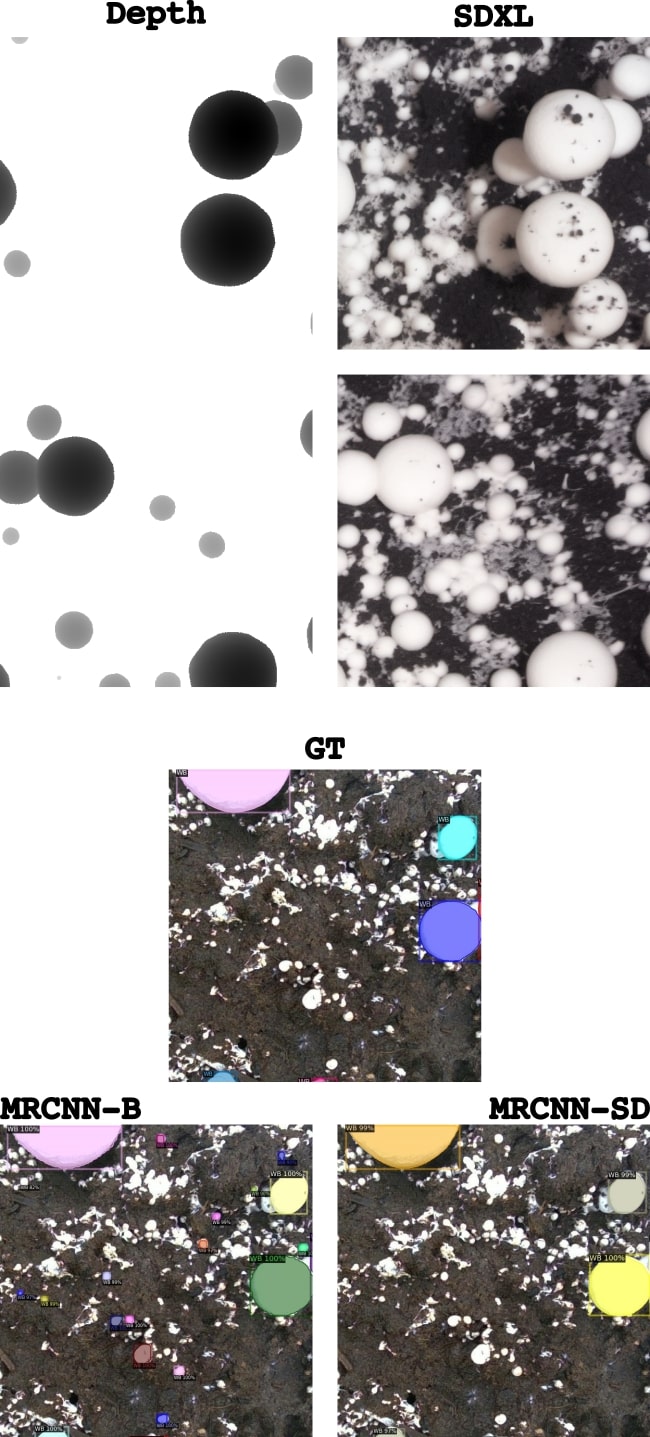}
        \caption{}
        \label{fig:subfig_7_b}
    \end{subfigure}
    \hfill
    \begin{subfigure}[b]{0.3\textwidth}
        \includegraphics[width=\textwidth]{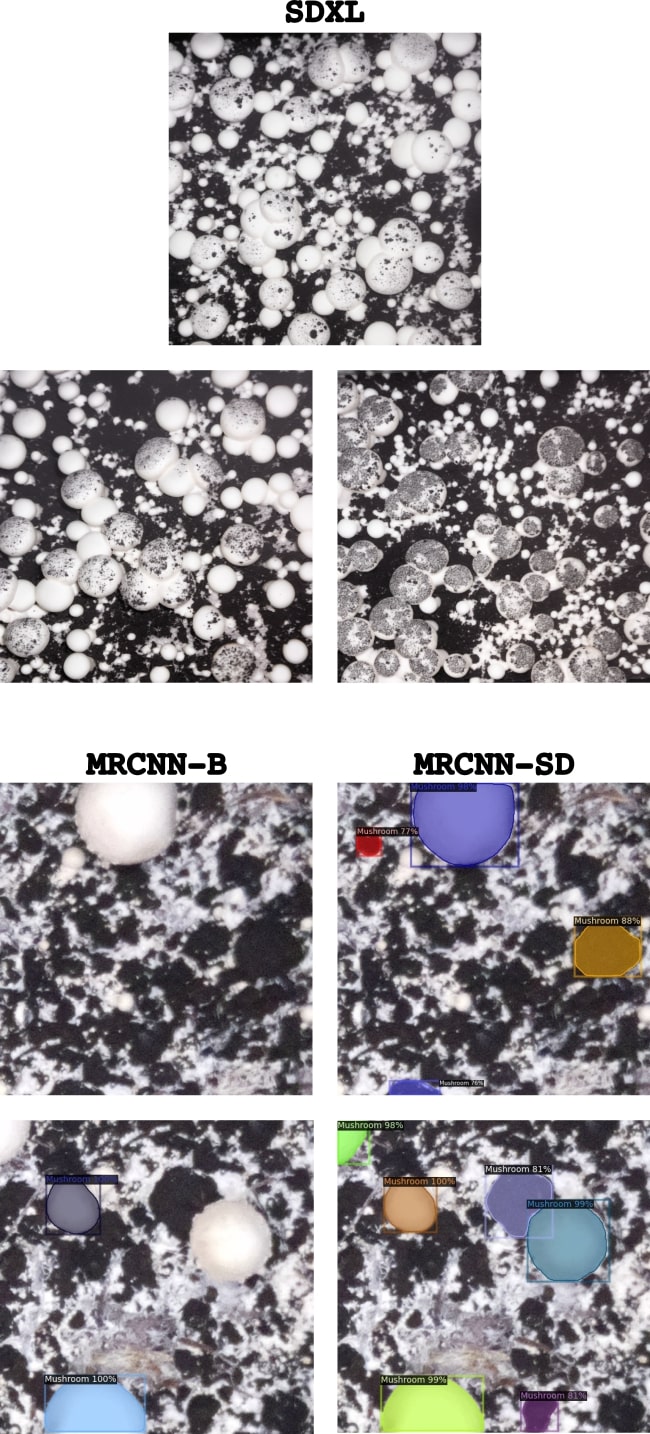}
        \caption{}
        \label{fig:subfig_7_c}
    \end{subfigure}
    \caption{Qualitative examples showcasing success and failure cases, as well as artifacts in image generation and segmentation. (a): Detection results for heavily occluded mushrooms with the MRCNN-B and MRCNN-SD models. (b): SDXL generating mushrooms where there shouldn't be any, according to the depth map, and comparison of MRCNN-B and MRCNN-SD on an M18K test image (MRCNN-B segments small mushrooms not present in the ground truth masks). (c): ``Dirty mushrooms'' generated by SDXL and its effect on MRCNN-SD false positives. }
    \label{fig:fig7}
\end{figure}

A closer examination of the generated images highlights both the strengths and limitations of our approach. In the Blender scene, randomized geometry results in an organically distributed mushroom layout, with clumping behavior typical of real-world conditions (Figures~\ref{fig:fig2}.~and \ref{fig:fig3}.) This clumping leads to frequent occlusions, particularly of smaller mushrooms by larger ones. As shown in Figures~\ref{fig:fig4}.~and \ref{fig:fig6}., ControlNet’s depth-based guidance remains robust across different random seeds and workflow configurations, consistently placing mushrooms at intended locations. Visual inspection confirms that this guidance allows segmentation masks from Blender to align accurately with SDXL-generated images. Moreover, the clumping behavior successfully transfers to the SDXL images, enabling models trained on these datasets to handle crowded scenes and to segment partially visible mushrooms effectively. Example results are shown in Figure~\ref{fig:fig7}/a.

While ControlNet reliably guides mushroom placement in intended regions, it occasionally permits unwanted mushrooms to appear in areas where none should be present. These extra mushrooms lack corresponding segmentation masks, which can hinder model training and affect performance. This issue was reflected in our experiments by the MRCNN-SD model’s higher precision but lower recall, indicating the model learned to ignore uncertain, small mushroom instances that lacked masks and instead focused on confidently detectable ones. Interestingly, this behavior gave MRCNN-SD an advantage over MRCNN-B on the M18K dataset, which also contains many small, unlabeled mushroom pins that MRCNN-SD effectively ignored. Examples of this effect, along with segmentation comparisons between MRCNN-B and MRCNN-SD on M18K, are shown in Figure~\ref{fig:fig7}/b. To address this challenge, Blender-generated segmentation masks could be incorporated into the SDXL workflow by either using a ControlNet variant that accepts segmentation masks or by applying inpainting, which uses masks to define regions where the diffusion model is allowed to generate content.

SDXL-generated images occasionally show unrealistic artifacts, such as excessive dirt on mushrooms or unusually dark mushroom colors, as seen in Figure~\ref{fig:fig7}/c. While the exact cause is unclear, we attribute it to the diffusion model’s tendency to favor detailed, high-contrast textures over uniform regions. Visual inspection of the ablation study results reveals that these artifacts are least prominent when all components of the proposed pipeline are utilized, suggesting that their combination helps mitigate such issues. Although this effect increases the domain gap between SDXL-generated and real images, it appears to have minimal impact on model performance, as indicated by the evaluation results. In rare cases, MRCNN-SD can produce false positives, misclassifying circular soil patches as mushrooms (see Figure~\ref{fig:fig7}/c). Such errors can be attributed to this effect.

\subsection{Stability and generalizability}

\begin{figure}[t]
    \centering

    \begin{subfigure}[b]{\textwidth}
        \centering
        \includegraphics[width=\textwidth]{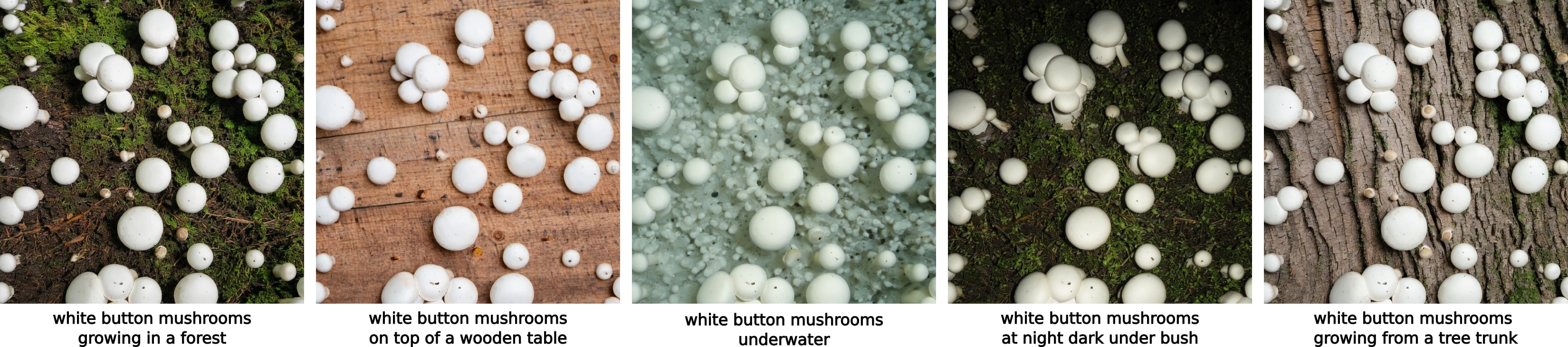}
        \caption{}
        \label{fig:subfig_8_a}
    \end{subfigure}

    \vspace{1em}

    \begin{subfigure}[b]{0.75\textwidth}
        \centering
        \includegraphics[width=0.75\textwidth]{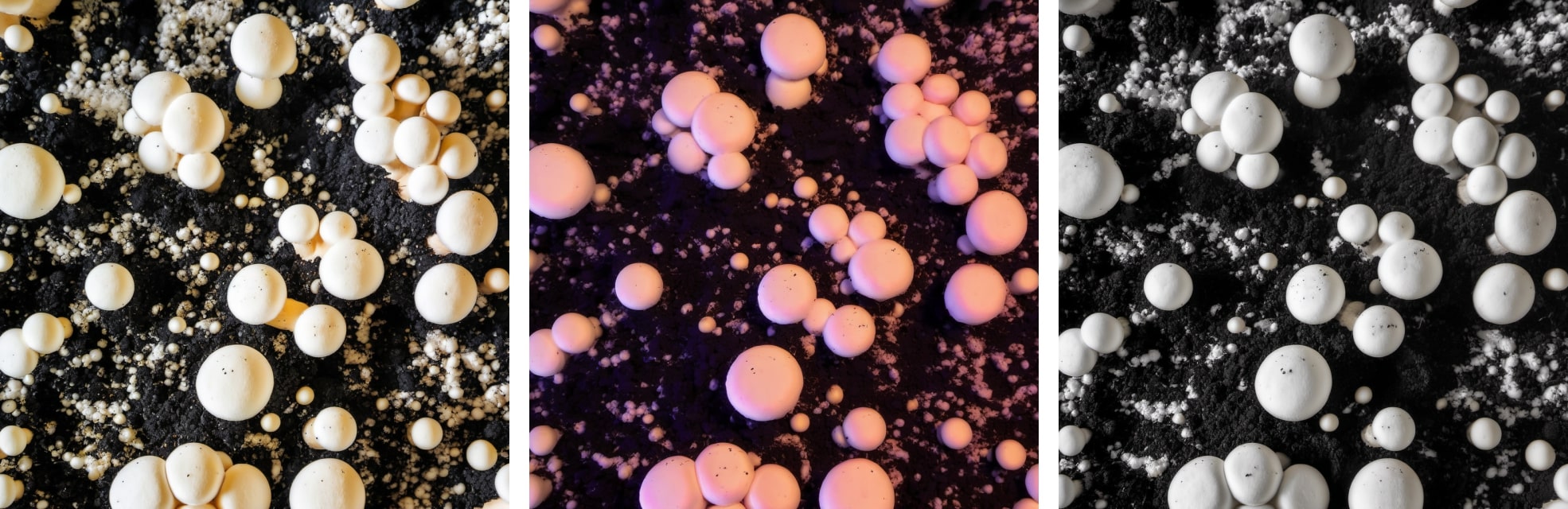}
        \caption{}
        \label{fig:subfig_8_b}
    \end{subfigure}

    \caption{Stability and generalizability of SDXL-generated images. (a): Images generated with the same random seed and depth map, without the IP-Adapter, using different text prompts. (b): Images generated with the same random seed, depth map, and default text prompt, using different image prompts.}
    \label{fig:fig8}
\end{figure}

As established earlier, the workflow demonstrates robustness to variations in the random seed. However, all previously shown SDXL-generated images in our synthetic dataset used the same text prompt: “white button mushrooms on dark black soil compost mycelium.” To demonstrate the stability of the workflow to prompting, we generated additional images using different text and image prompts while keeping the depth map and random seeds fixed. The results, shown in Figure~\ref{fig:fig8}., reveal that significant domain shifts can be achieved by altering the text prompt without affecting the consistency of mushroom placement or appearance. More subtle stylistic variations, such as lighting and camera effects, can be introduced through different image prompts. These results suggest that the proposed workflow is both flexible and robust to domain shifts.

The proposed workflow can be easily adapted to other domains, such as fruit or leaf segmentation. The setup process involves:

\begin{enumerate}
    \item Creating a Blender scene that accurately reflects the geometry of the target domain (e.g., modeling fruits or leaves, and relevant scene elements like branches), and configuring it with BAT for automatic generation of segmentation masks and depth maps.
    \item Collecting a small set (15--20) of reference images from the web or on-site for each key visual component in the given domain.
    \item Annotating the collected reference images with descriptive text prompts.
    \item Training LoRAs on the annotated reference images for key visual components.
    \item Selecting reference images for the IP-Adapter that best match the real-world style and lighting conditions.
    \item Using the full workflow to synthesize a domain-specific dataset.
\end{enumerate}

\section{Conclusion}
\label{sec:conc}

This paper introduced a novel synthetic data generation pipeline for white button mushroom segmentation that combines the precise control of 3D computer graphics provided by Blender with the photorealistic capabilities of Stable Diffusion XL-based text-to-image diffusion models. By guiding image generation with depth maps extracted from the synthetic scene, this hybrid workflow overcomes the need for manual configuration of complex rendering parameters such as shaders, materials, lighting, and post-processing typically required for photorealistic outputs.

Our approach enables rapid creation of high-quality, annotated synthetic datasets, significantly reducing dataset preparation effort. Evaluations demonstrate that models trained exclusively on data generated through this pipeline achieve segmentation performance comparable to those trained on real-world images, as validated on the M18K benchmark dataset. Additionally, ablation studies highlight the importance of key components — including ControlNet, IP-Adapter, and LoRA modules — in enhancing image fidelity and domain alignment.

While certain limitations remain, such as occasional generation artifacts and incomplete suppression of unwanted mushrooms in generated images, our results confirm the robustness, generalizability, and practical utility of the proposed pipeline. The methodology is broadly applicable beyond white button mushrooms and can be easily adapted to other agricultural and industrial domains requiring scalable training data generation for machine learning applications.

\section*{Data availability}

Data is available on our Hugging Face page \cite{synwmb} (\href{https://doi.org/10.57967/hf/6084}{\url{https://doi.org/10.57967/hf/6084}}). Usage examples and complementary code are also hosted in this repository.

\section*{Abbreviations}
The following abbreviations are used in this manuscript:

\noindent 
\begin{tabular}{@{}ll}
AP & Average Precision \\
AR & Average Recall \\
BAT & Blender Annotation Tool \\
CUT &  Contrastive learning for Unpaired image-to-image Translation \\
DBSCAN & Density-Based Spatial Clustering of Applications with Noise \\
DL & Deep Learning \\
FCGF & Fully Convolutional Geometric Feature \\
FID & Fréchet Inception Distance \\
FPFH & Fast Point Feature Histogram \\
GAN & Generative Adversarial Network \\
GT & ground-truth \\
HDRI & High Dynamic Range Image \\
IP-Adapter & Image Prompt Adapter \\
KID & Kernel Inception Distance \\
LoRA & Low-Rank Adaptation \\
mIoU & mean Intersection over Union \\
MRCNN-B & Mask R-CNN model trained on Blender-rendered data \\
MRCNN-SD & Mask R-CNN model trained on Stable Diffusion data \\
MSE & Mean Squared Error \\
PNDR & Photo-realistic Neural Domain Randomization \\
SDXL & Stable Diffusion XL \\
SSD & Single Shot Detector
\end{tabular}

\section*{Acknowledgment}

This research was supported by the Consolidator Researcher Grant of Obuda University (Grant ID: OKPPKPC004). Artúr I.~Károly is supported by the 2024-2.1.1 University Research Scholarship Program of the Ministry for Culture and Innovation from the source of the National Research, Development and Innovation Fund.

\section*{Contributions}
A.I.K.: Conceptualization, Methodology, Software, Validation, Formal analysis, Investigation, Data curation, Writing---original draft, Writing---review and editing, Visualization, Project administration, Funding acquisition

\noindent
P.G.: Conceptualization, Resources, Writing---review and editing, Supervision, Project administration, Funding acquisition. All authors have read and agreed to the published version of the manuscript.

\end{document}